\documentclass[review,3p,10pt]{elsarticle}

\usepackage{lineno,hyperref}
\usepackage{multicol}
\usepackage{multirow}
\usepackage{amsmath}
\usepackage{wrapfig}
\usepackage{threeparttable}
\usepackage{flushend}

\journal{Neurocomputing}

\begin{document}

\begin{frontmatter}



\title{A Heterogeneous Branch and Multi-Level Classification Network for Person Re-Identification}

\author[mymainaddress]{Jiabao Wang\corref{correspondingauthor}}
\ead{jiabao\_1108@163.com}
\cortext[correspondingauthor]{Corresponding author}

\author[mymainaddress]{Yang Li}
\ead{solarleeon@outlook.com}

\author[mymainaddress]{Yangshuo Zhang}
\ead{17625944869@163.com}

\author[mymainaddress]{Zhuang Miao}
\ead{emiao\_beyond@163.com}

\author[mymainaddress]{Rui Zhang}
\ead{3959966@qq.com}

\address[mymainaddress]{College of Command and Control Engineering, Army Engineering University of PLA, China}

\begin{abstract}
Convolutional neural networks with multiple branches have recently been proved highly effective in person re-identification (re-ID). Researchers design multi-branch networks using part models, yet they always attribute the effectiveness to multiple parts. In addition, existing multi-branch networks always have isomorphic branches, which lack structural diversity. In order to improve this problem, we propose a novel Heterogeneous Branch and Multi-level Classification Network (HBMCN), which is designed based on the pre-trained ResNet-50 model. A new heterogeneous branch, SE-Res-Branch, is proposed based on the SE-Res module, which consists of the Squeeze-and-Excitation block and the residual block. Furthermore, a new multi-level classification joint objective function is proposed for the supervised learning of HBMCN, whereby multi-level features are extracted from multiple high-level layers and concatenated to represent a person. Based on three public person re-ID benchmarks (Market1501, DukeMTMC-reID and CUHK03), experimental results show that the proposed HBMCN reaches 94.4\%, 85.7\% and 73.8\% in Rank-1, and 85.7\%, 74.6\% and 69.0\% in mAP, achieving a state-of-the-art performance. Further analysis demonstrates that the specially designed heterogeneous branch performs better than an isomorphic branch, and multi-level classification provides more discriminative features compared to single-level classification. As a result, HBMCN provides substantial further improvements in person re-ID tasks.
\end{abstract}


\end{frontmatter}


\section{Introduction}
\label{intro}
Pedestrian re-identification (re-ID) is a key technological process for the identification of the same pedestrian across different cameras. Not only is it is a basic task of intelligent surveillance, but it is also very important for public security. With the development of deep learning techniques, person re-ID has attracted much attention, achieving considerable improvements in terms of accuracy. Recently, Convolutional Neural Networks (CNNs) with multiple branches have demonstrated a surprising performance in person re-ID. This method divides the whole body into multiple parts, and designs networks with a global branch and multiple part branches. Yao et al.~\cite{DBLP:journals/corr/YaoZZLT17} proposed the Part Loss Networks (PL-Net), which automatically detects human body parts and computes a classification loss on each part. Sun et al.~\cite{DBLP:conf/eccv/SunZYTW18} proposed a Part-based Convolutional Baseline with a Refined Part Pooling (PCB-RPP) method, which introduces multiple branches for training networks. Wang et al.~\cite{DBLP:conf/mm/WangYCLZ18} designed a Multiple Granularity Network (MGN), with one global branch and two local branches. These part-based models use multiple branches to achieve state-of-the-art performance. However, researchers always attribute the effectiveness to multiple parts, not multiple branches. Thus, in this paper we explore the contribution of multiple branches, without using part branches. Furthermore, the existing multi-branch networks use isomorphic branches, which are replicated from the high-level layers of the global branch. As a result, there is a lack of structural diversity for learning differentiated features.

In this paper, we propose a novel Heterogeneous Branch and Multi-level Classification Network (HBMCN). The contributions of this work are as follows:

a) A novel heterogeneous branch and multi-level classification network is proposed for person re-ID. It is designed based on the ImageNet pre-trained ResNet-50 model and consists of heterogeneous branches. Multiple softmax log-loss objectives are set for supervised learning, at multiple high-level layers of HBMCN.

b) A new heterogeneous branch, SE-Res-Branch, is proposed to enhance the structural diversity of HBMCN. SE-Res-Branch is designed based on the SE-Res module, which consists of the Squeeze-and-Excitation (SE) block and the residual block. The SE block adaptively recalibrates channel-wise feature responses of the residual block, by explicitly modeling interdependencies between channels.

c) A new multi-level classification joint objective function is proposed to learn the parameters of HBMCN. For each branch, multiple classification objectives are designed at multiple high-level layers and produce multi-level features. The features are concatenated to form a final feature vector for person representation.

d) To evaluate HBMCN, we explore the effect of the branch type, multi-level feature, and so on. Experimental results show that the specially designed heterogeneous branch performs better than an isomorphic branch, and multi-level classification provides more discriminative features compared to single-level classification.

\section{Related Work}
Person re-ID has recently attracted a lot of attention due to public security needs. Much of the existing work focuses on feature learning and metric learning.

Prior to the popularity of deep learning, much research was performed in order to explore design hand-crafted features, such as Local Binary Pattern (LBP)~\cite{DBLP:conf/eccv/XiongGCS14} features and Local Maximal Occurrence (LOMO)~\cite{DBLP:conf/cvpr/LiaoHZL15} features. With the development of deep learning, deep feature representation became the dominant method, making significant progress in person re-ID. Currently, in the deep learning age, researchers pay more attention to design networks for learning features. In traditional methods, researchers always divide a person image into multiple parts, and this idea is also transferred into deep learning methods. For part-based deep learning methods, such as PL-Net~\cite{DBLP:journals/corr/YaoZZLT17}, PCB-RPP~\cite{DBLP:conf/eccv/SunZYTW18} and MGN~\cite{DBLP:conf/mm/WangYCLZ18}, a global branch is used for global information and multiple part branches are used for local information. The effect of part-based models can be achieved by using either multiple parts, multiple branches, or both. However, researchers always attribute the effectiveness to multiple parts, ignoring the contribution of multiple branches. Thus, in this paper we explore the effect of multiple branches, without the use of local part information. Existing multi-branch networks use isomorphic branches, which lack the structural diversity required for learning differentiated features. As a result, we explore a new heterogeneous branch to enhance the capability of person re-ID. For heterogeneous branches, the PAN~\cite{DBLP:journals/corr/ZhengZY17aa} method uses a base branch and an alignment branch for simultaneously aligning pedestrians within images and learning pedestrian descriptors. In addition, Zhao et al.~\cite{DBLP:conf/iccv/ZhaoLZW17} proposed the deeply-learned part-aligned representation for re-ID. Both also use different branches for person re-ID, yet their aim is to align the deformable parts of the person. However, our heterogeneous branch is directly used for learning differential features.

In terms of person re-ID, metric learning~\cite{DBLP:journals/corr/YiLL14,DBLP:conf/eccv/ShiYZLLZL16} is an additional research direction for similarity ranking. The fundamental idea of this method is to treat the person re-ID as a supervised distance metric learning problem. Traditional methods, such as Keep It Simple and Straight forward Metric (KISSME)~\cite{DBLP:conf/cvpr/KostingerHWRB12} and Cross-view Quadratic Discriminant Analysis (XQDA)~\cite{DBLP:conf/cvpr/LiaoHZL15}, try to learn a transform matrix of features. For deep learning methods, metric learning is introduced in the form of the contractive loss~\cite{DBLP:journals/corr/ZhengZY16}, the triplet loss~\cite{DBLP:journals/corr/HermansBL17}, and quadratic loss~\cite{DBLP:conf/cvpr/ChenCZH17}. In addition, specially designed losses, such as the cosine loss~\cite{DBLP:conf/cccv/WangLM17} and sphere loss~\cite{DBLP:journals/corr/abs-1807-00537}, are also proposed for improved feature representation learning. In practice, the combination of multiple losses can reach a better performance. For example, MGN~\cite{DBLP:conf/mm/WangYCLZ18} combines classification loss and triplet loss to achieve a surprising performance (see Table \ref{tab_state-of-the-arts} in Section \ref{sec_comparison}). However, in this paper, we pay more attention to the heterogeneous branch, and use the simplest softmax log-loss for each branch in our network.

ResNet is a widely used model in person re-identification ~\cite{DBLP:journals/corr/ZhengYH16,DBLP:journals/corr/ZhengZY16,DBLP:conf/cccv/WangLM17}, whereby the residual block is designed to tackle the gradient vanishing problem. ResNet has many paths from the input to the output with skip connections and can be treated as a kind of ensemble model~\cite{DBLP:conf/cvpr/HeZRS16}. However, the network has only one loss, and the parameters in the multiple paths are shared completely. The learned parameters just minimize one softmax log-loss and there is a lack of diversity. GoogleNet is a very successful model with multiple objectives for a great performance~\cite{DBLP:conf/cvpr/SzegedyLJSRAEVR15}. However, their multiple objectives with different lengths are also designed for tackling the gradient vanishing problem. Compared with ResNet, the depth of GoogleNet is shallow and limits its generalization ability. Motivated by the idea of multiple objectives with different lengths, we design a multi-level classification objective function based on the ResNet model, and extract features before each of the objectives for person representation.

\section{HBMCN}
\subsection{Architecture}
\begin{figure*}[!t]
\centering
\includegraphics[width=4.8in]{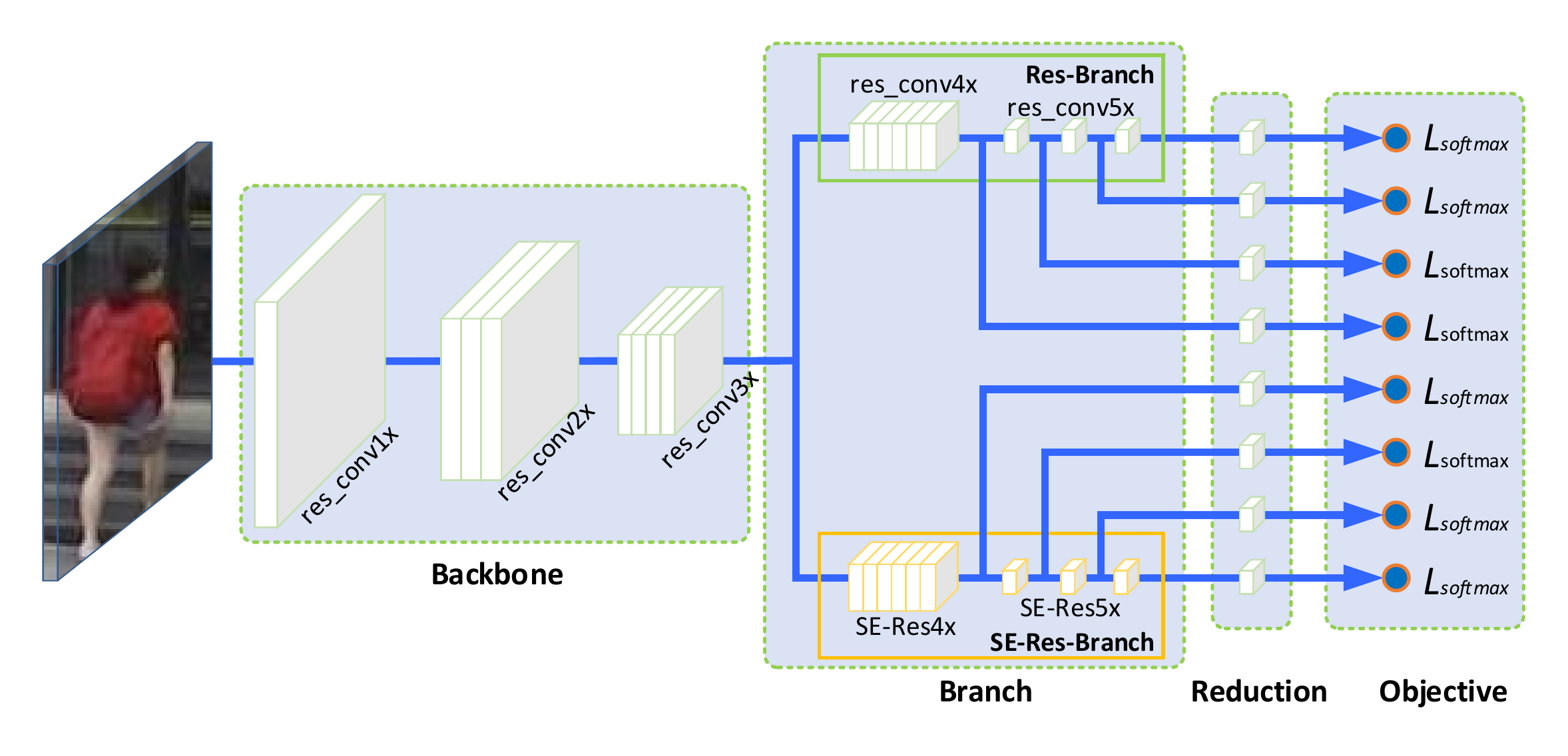}
\caption{Architecture of the proposed HBMCN. The main features include a parameters-shared \emph{Backbone}, two parameter-independent heterogeneous \emph{Branches}, multiple \emph{Reductions} and \emph{Objectives} for feature extraction and supervised learning. The specially designed \emph{SE-Res-Branch} can enhance the structural diversity, while the multi-level classification joint objective function can produce multi-level discriminative features.}
\label{fig_architecture}
\end{figure*}

Figure \ref{fig_architecture} shows the architecture of our proposed Heterogeneous Branch and Multi-level Classification Network (HBMCN). It is based on the ImageNet pre-trained ResNet-50 model~\cite{DBLP:conf/cvpr/HeZRS16} and can be divided into four parts: \emph{Backbone}, \emph{Branch}, \emph{Reduction} and \emph{Objective}. The first part is the parameters-shared backbone sub-network, and the second part is the special designed parameter-independent heterogeneous branch. The third part is the dimension reduction for feature extraction. The last softmax log-loss objectives are used for classification. The whole network has several branches for feature extraction.

\paragraph{Backbone} Based on ResNet-50, the backbone of HBMCN includes the layers of \emph{res\_conv1x}, \emph{res\_conv2x} and \emph{res\_conv3x}. The backbone shares the parameters and computation, and can input a person image and output a set of low-level feature maps. After the backbone, the network breaks into two heterogeneous branches.

\paragraph{Branch} This consists of two heterogeneous branches: \emph{Res-Branch} and \emph{SE-Res-Branch}. \emph{Res-Branch} is composed of the \emph{res\_conv4x} and  \emph{res\_conv5x} layers of ResNet-50. \emph{SE-Res-Branch} uses the \emph{SE-Res4x} and \emph{SE-Res5x} layers correspondingly. It is designed based on the SE-Res module, which consists of the Squeeze-and-Excitation (SE) block\cite{DBLP:conf/cvpr/HuSS18} and the residual block. Details of \emph{SE-Res-Branch} are presented in section \ref{se-res-branch}.

\paragraph{Reduction} The output of the \emph{res\_conv5x} (\emph{res\_conv4x}) layer is a 2048-dims (1024-dims) feature following the Global Average Pooling (GAP) operation. To represent a person effectively, we decrease the high-dimensional feature. \emph{Reduction} is composed of a $1 \times 1$ convolution, followed by a batch normalization and a leaky ReLU with the negative slope of 0.1. This reduces the number of channels from 2048-dims(1024-dims) to 256-dims, which is just $1/8$($1/4$) of the original dimension.

\paragraph{Objective} The multi-level classification joint objective function is set at multiple high-level layers of the two branches. It derives from the outputs of \emph{res\_conv4f}, \emph{res\_conv5a}, \emph{res\_conv5b}, \emph{res\_conv5c} \emph{SE-Res4f}, \emph{SE-Res5a}, \emph{SE-Res5b}, and \emph{SE-Res5c}. For each feature produced by \emph{Reduction}, we use a $1 \times 1 $ convolution to map the 256-dims feature to the number of person identities. The eight softmax log-loss objectives are jointly used for classification.

\subsection{SE-Res-Branch}
\label{se-res-branch}
\begin{figure}[!t]
\centering
\includegraphics[width=3.5in]{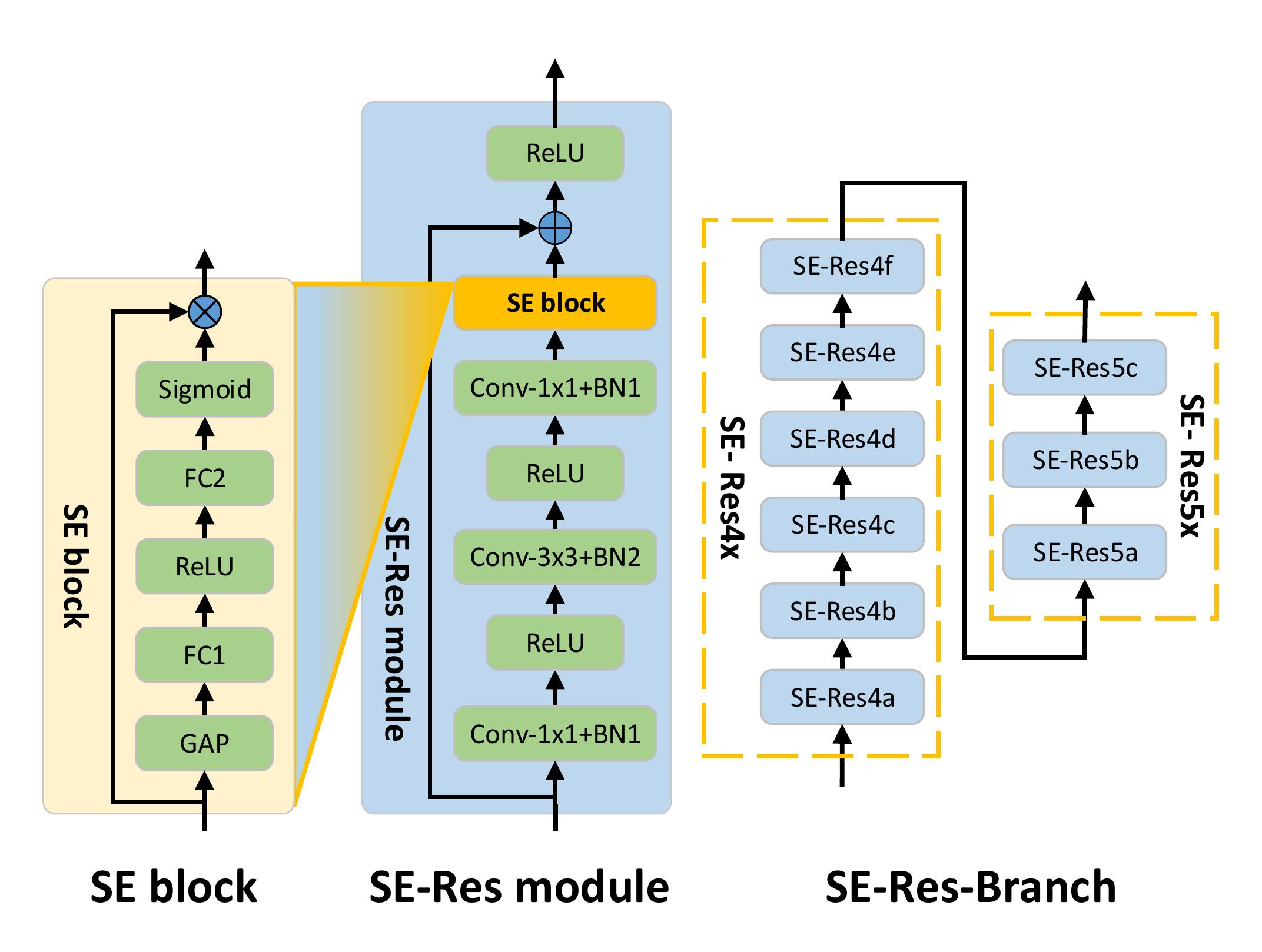}
\caption{SE-Res-Branch, consisting of a stack of the SE-Res modules, which can be further divided into the Squeeze-and-Excitation (SE) block and the residual block.}
\label{fig_se-residual-branch}
\end{figure}

As shown in Figure \ref{fig_se-residual-branch}, \emph{SE-Res-Branch} is based on the SE-Res module, which consists of the Squeeze-and-Excitation (SE) block\cite{DBLP:conf/cvpr/HuSS18} and the residual block~\cite{DBLP:conf/cvpr/HeZRS16}.

The \textbf{SE block} in Figure \ref{fig_se-residual-branch} is built with \emph{squeeze} and \emph{excitation} units to perform feature recalibration.

The \emph{squeeze} unit is used to squeeze global spatial information into a channel descriptor. It can be achieved using GAP to generate channel-wise statistics. For an input tensor $\bf{Y}=[\bf{y}_1,\bf{y}_2,...,\bf{y}_C]\in {R}^{H\times W \times C}$, the result ${\bf{z} \in R^C}$ is generated by shrinking $\bf{Y}$ through the spatial dimensions $H\times W$, and the $c$-th element of $\bf{z}$ is calculated using
\begin{equation}\label{eq1}
\begin{aligned}
{{\bf{z}}_c} &= {{\bf{F}}_{sq}}({{\bf{y}}_c})\\
 &= \frac{1}{{H \times W}}\sum\limits_{i = 1}^H {\sum\limits_{j = 1}^W {{{\bf{y}}_c}(i,j)} },
\end{aligned}
\end{equation}
where ${\bf{F}}_{sq}$ refers to the \emph{squeeze} function.

The \emph{excitation} unit is used to produce a channel-wise weight vector. The output $\bf{s} \in {R}^C$ is obtained by a series of transformations with a sigmoid activation as follows:
\begin{equation}\label{eq2}
\begin{aligned}
{\bf{s}} &= {{\bf{F}}_{ex}}({\bf{z}},{\bf{W^{se}}}) \\
&= \sigma ({{\bf{W}}_2^{se}}\delta ({{\bf{W}}_1^{se}}{\bf{z}})),
\end{aligned}
\end{equation}
where ${\bf{F}}_{ex}$ refers to the \emph{excitation} function, $\sigma$ refers to the sigmoid function and $\delta$ refers to the ReLU function. The term $\bf{W}_1^{se}$ is a linear transformation parameter used for dimensionality-reduction, and $\bf{W}_2^{se}$ is a linear transformation parameter for dimensionality-increasing.

The output of the SE block is computed as:
\begin{equation}\label{eq3}
\begin{aligned}
{{\bf{y'}}_c} &= {{\bf{F}}_{scale}}({{\bf{y}}_c},{s_c})\\
&= {{\bf{y}}_c} \cdot {s_c},
\end{aligned}
\end{equation}
where ${\bf{F}}_{scale}$ refers to the rescaling function, which is a channel-wise multiplication between the feature map $\bf{y}_c \in {R}^{H\times W}$ and the scalar $s_c$.

\textbf{SE-Res module} consists of the residual block and the SE block. The residual block uses a stack of 3 ($1\times 1$, $3\times 3$ and $1\times 1$) convolutional layers, where the two $1\times 1$ layers are responsible for reducing and subsequently increasing dimensions, leaving the $3\times 3$ layer as a bottleneck with smaller input and output dimensions. For an input tensor $\bf{X} \in {R}^{H\times W \times C}$, the output $\bf{Y}$ can be computed as
\begin{equation}\label{eq4}
\begin{aligned}
{\bf{Y}} &= {{\bf{F}}_{res}}({\bf{X}},{{\bf{W}}^{res}}) \\
&= {\bf{W}}_3^{res}\odot \delta ({\bf{W}}_2^{res}\odot \delta ({\bf{W}}_1^{res}\odot{\bf{X}})),
\end{aligned}
\end{equation}
where ${\bf{F}}_{res}$ refers the residual function, $\odot$ denotes convolution, and $\bf{W}_1^{res}$ , $\bf{W}_2^{res}$ and $\bf{W}_3^{res}$ are the parameters of the $1\times 1$, $3\times 3$ and $1\times 1$ convolutional layers. $\bf{Y}$ is the input of SE block, and the corresponding $\bf{Y'}=[\bf{y'}_1,\bf{y'}_2,...,\bf{y'}_C]$ is the output of SE block.

The whole output of the SE-Res module can be computed as
\begin{equation}\label{eq5}
{{\bf{o}}_c} = \delta ({{\bf{y'}}_c} + {{\bf{x}}_c}),
\end{equation}
where ${\bf{O}} = [{{\bf{o}}_1},{{\bf{o}}_2},...,{{\bf{o}}_C}]$, and ${\bf{X}} = [{{\bf{x}}_1},{{\bf{x}}_2},...,{{\bf{x}}_C}]$.

\textbf{SE-Res-Branch} is designed based on the SE-Res modules. According to the \emph{res\_conv4x} and \emph{res\_conv5x} layers of ResNet-50, we design the equivalent \emph{SE-Res4x} and \emph{SE-Res5x} layers, where there are 6 and 3 SE-Res modules respectively. The first module of \emph{SE-Res4x} and \emph{SE-Res5x}  has a $1 \times 1 $ convolution with a stride 2 to reduce the resolution of the feature maps.

\subsection{Multi-Level Classification}
For the part-based models, such as PCB-RPP~\cite{DBLP:conf/eccv/SunZYTW18} and MGN~\cite{DBLP:conf/mm/WangYCLZ18}, the features are computed from the output of \emph{res\_conv5c}, which is the last residual block in ResNet-50. Inspired by GoogleNet~\cite{DBLP:conf/cvpr/SzegedyLJSRAEVR15}, we set multiple objectives from the layers of \emph{res\_conv4f}, \emph{res\_conv5a}, \emph{res\_conv5b} and \emph{res\_conv5c} and the layers of \emph{SE-Res4f}, \emph{SE-Res5a}, \emph{SE-Res5b} and \emph{SE-Res5c}. Before each objective, we adopt \emph{Reduction} to reduce the dimension of the original feature. The dimension of \emph{res\_conv4f} and \emph{SE-Res4f} is 1024, while the dimension of the remaining is 2048. Thus, the $1\times 1$ convolution of \emph{Reduction} is used to reduce the different dimension to the same value of 256. Following \emph{Reduction}, eight features can be obtained from the multiple high-level layers.

Given a batch of person images ${\{I_i\}}_{i=1}^B$, for the $k$-th objective, the corresponding feature $\mathbf{f}^k(I_i)$ can be obtained from the $k$-th \emph{Reduction}. The softmax log-loss is then computed from feature $\mathbf{f}^k(I_i)$ and its truth label ${y}_{i}$. Each objective corresponds to one loss, which has the form of
\begin{equation}\label{eq6}
{{L}_{softmax}^k}=-\sum\limits_{i=1}^{B}{\log \frac{\exp ((\mathbf{W}_{{{y}_{i}}}^k)^{T}{\mathbf{f}^k(I_i)}+{{b}_{{{y}_{i}}}^k})}{\sum\nolimits_{j=1}^{C}{\exp ((\mathbf{W}_{j}^k)^{T}{\mathbf{f}^k(I_i)}+{{b}_{j}}^k)}}},
\end{equation}
where $B$ is the mini-batch size, $C$ is the number of classes, and $\mathbf{W}_{j}^k$ and $b_{j}^k$ are the parameters of the $k$-th objective to learn.

The multi-level classification joint objective is the summation of all eight objectives,
\begin{equation}\label{eq7}
{L_{joint}} = \sum\limits_{k = 1}^8 {L_{softmax }^k}.
\end{equation}

After training, given a query or gallery person image $I_t$, its representation can be obtained by concatenating the eight features $\mathbf{f}^k(I_t), k=1,2,...,8$
\begin{equation}\label{eq8}
{\bf{f}}(I_t) = [{{\bf{f}}^1}(I_t),{{\bf{f}}^2}(I_t),...,{{\bf{f}}^8}(I_t)]
\end{equation}

In practice, the features can also be directly extracted from the layers of \emph{res\_conv4f}, \emph{res\_conv5a}, and \emph{res\_conv5b} and those of \emph{SE-Res4f}, \emph{SE-Res5a}, \emph{SE-Res5b}, without \emph{Reduction} and \emph{Objective}. However, the features produced by GAP from these layers have 2048(1024) dimensions, which is too high for integration. On the contrary, our multiple independent objectives can produce low-dimensional features by \emph{Reduction}. In addition, multiple objectives can supervise the network to learn more discriminative features.

\section{Experiments}
In order to evaluate the effectiveness of our HBMCN, we perform experiments on three public datasets, including Market1501~\cite{DBLP:conf/iccv/ZhengSTWWT15}, DukeMTMC-reID~\cite{DBLP:conf/iccv/ZhengZY17} and CUHK03~\cite{DBLP:conf/cvpr/LiZXW14}. In all experiments, we abbreviate Market1501, DukeMTMC-reID and CUHK03 to Market, Duke, and CUHK. Market contains 751 identities of 12936 images for training, and 750 identities, clutter and background for testing. Duke is a subset of the DukeMTMC, and consists of 36411 images of 1404 identities. CUHK contains 14097 images of 1467 identities captured from cameras at the Chinese University of Hong Kong campus. The statistics of the datasets are presented in Table \ref{tab_datasets}. For Market and Duke, we use the standard evaluation protocol~\cite{DBLP:conf/iccv/ZhengSTWWT15}, while the new training and testing protocol is used for CUHK~\cite{DBLP:conf/cvpr/ZhongZCL17}.

\begin{table*}[]
\centering
\caption{Experimental datasets.}
\label{tab_datasets}
\begin{threeparttable}
\begin{tabular}{|l|c|c|c|c|c|c|c|c|}
\hline
\multirow{2}{*}{\textbf{Datasets}} & \multicolumn{2}{c|}{\textbf{train}} & \multicolumn{2}{c|}{\textbf{gallery}} & \multicolumn{2}{c|}{\textbf{query}} & \multicolumn{2}{c|}{\textbf{total}} \\ \cline{2-9}
 & \textbf{images} & \textbf{ids} & \textbf{images} & \textbf{ids} & \textbf{images} & \textbf{ids} & \textbf{image} & \textbf{ids} \\ \hline
\textbf{Market} & 12936 & 751 & 19732 & 750* & 3368 & 750 & 36036 & 1501 \\ \hline
\textbf{Duke} & 16522 & 702 & 17661 & 1110 & 2228 & 702 & 36411 & 1404 \\ \hline
\textbf{CUHK} & 7365 & 767 & 5332 & 700 & 1400 & 700 & 14097 & 1467 \\ \hline
\end{tabular}
\begin{tablenotes}
    \footnotesize
    \item[1] `*' means that the number doesn't include the background and junks.
\end{tablenotes}
\end{threeparttable}
\end{table*}

\subsection{Implementation}
For the implementation stage, we initialize the parameters of the \emph{Backbone} and \emph{Res-Branch} by the parameters of the ImageNet pre-trained ResNet-50. \emph{SE-Res-Branch} and additional network parameters are initialized by the `arxiver' method~\cite{DBLP:conf/cvpr/HeZRS16}. The input image is uniformly resized to $384 \times 192$.

For training, the data augmentation is adopted. This includes random cropping, horizontal flipping and random erasing~\cite{DBLP:journals/corr/abs-1708-04896}. The mini-batch size of training is 32, and the examples are shuffled randomly. The SGD optimizer is used with a momentum of 0.9. The weight decay factor is set to 0.0005. The learning rate is initialized from 0.01, and the decay to 0.001 and 0.0001, after training for 40 and 60 epochs, respectively. The total training has 80 epochs. The learning rate of \emph{SE-Res-Branch}, \emph{Reduction} and \emph{Objective} modules is 10 times that of the learning rate of the pre-trained parameters.

For testing, we average the features extracted from an original image and its horizontal flipped version as the final feature. The cosine similarity is used for evaluation. Our model is implemented on the Pytorch v0.4 framework. The training of the Market1501 dataset with one NVIDIA GTX 1080Ti GPU lasts approximately 3 hours. To compare the performance of different methods, the two public evaluation metrics, CMC and mean Average Precision (mAP), are used. In all experiments, we use the single query mode and report the CMC at rank-1, rank-5, rank-10 and rank-20, and mAP~\cite{DBLP:conf/iccv/ZhengSTWWT15}.

\subsection{Comparison with state-of-the-art methods}
\label{sec_comparison}
To test the performance of HBMCN, we compare it with state-of-the-art methods, such as IDE model~\cite{DBLP:journals/corr/ZhengYH16}, PAN~\cite{DBLP:journals/corr/ZhengZY17aa}, SVDNet~\cite{DBLP:conf/iccv/SunZDW17}, TriNet~\cite{DBLP:journals/corr/HermansBL17}, DaRe~\cite{DBLP:journals/corr/abs-1805-08805}, SAG~\cite{DBLP:journals/corr/abs-1809-08556}, MLFN~\cite{DBLP:journals/corr/abs-1803-09132}, HA-CNN~\cite{DBLP:journals/corr/abs-1802-08122}, DuATM~\cite{DBLP:journals/corr/abs-1803-09937}, DeepPerson~\cite{DBLP:conf/icb/JinWLL17}, Fusion~\cite{DBLP:journals/corr/abs-1803-10630}, SphereReID~\cite{DBLP:journals/corr/abs-1807-00537}, PL-Net~\cite{DBLP:journals/corr/YaoZZLT17}, PCB~\cite{DBLP:conf/eccv/SunZYTW18} and MGN~\cite{DBLP:conf/mm/WangYCLZ18}. Detailed results are reported in Table \ref{tab_state-of-the-arts}, where the results of part-models, PL-Net~\cite{DBLP:journals/corr/YaoZZLT17}, PCB~\cite{DBLP:conf/eccv/SunZYTW18} and MGN~\cite{DBLP:conf/mm/WangYCLZ18}, are shown separately from other methods.

\begin{table*}[!t]
\centering
\caption{Comparison with state-of-the-art methods.}
\label{tab_state-of-the-arts}
\begin{tabular}{|l|c|c|c|c|c|c|}
\hline
 & \multicolumn{2}{c|}{\textbf{Market}} & \multicolumn{2}{c|}{\textbf{Duke}} & \multicolumn{2}{c|}{\textbf{CUHK}} \\ \hline
\textbf{Methods} & \textbf{mAP} & \textbf{Rank1} & \textbf{mAP} & \textbf{Rank1} & \textbf{mAP} & \textbf{Rank1} \\ \hline
\textbf{IDE~\cite{DBLP:journals/corr/ZhengYH16}} & 50.7\% & 75.6\% & 45.0\% & 65.2\% & 19.7\% & 21.3\% \\
\textbf{PAN~\cite{DBLP:journals/corr/ZhengZY17aa}} & 63.4\% & 82.8\% & 51.5\% & 71.6\% & 34.0\% & 36.3\% \\
\textbf{SVDNet~\cite{DBLP:conf/iccv/SunZDW17}} & 62.1\% & 82.3\% & 56.8\% & 76.7\% & 37.3\% & 41.5\% \\
\textbf{TriNet~\cite{DBLP:journals/corr/HermansBL17}} & 69.1\% & 84.9\% & -- & -- & 50.7\% & 55.5\% \\
\textbf{DaRe(R)~\cite{DBLP:journals/corr/abs-1805-08805}} & 69.3\% & 86.4\% & 57.4\% & 75.2\% & 51.3\% & 55.1\% \\
\textbf{DaRe(De)~\cite{DBLP:journals/corr/abs-1805-08805}} & 69.9\% & 86.0\% & 56.3\% & 74.5\% & 50.1\% & 54.3\% \\
\textbf{SAG~\cite{DBLP:journals/corr/abs-1809-08556}} & 73.9\% & 90.2\% & 60.9\% & 79.9\% & -- & -- \\
\textbf{MLFN~\cite{DBLP:journals/corr/abs-1803-09132}} & 74.3\% & 90.0\% & 62.8\% & 81.0\% & 47.8\% & 52.8\% \\
\textbf{HA-CNN~\cite{DBLP:journals/corr/abs-1802-08122}} & 75.5\% & 91.2\% & 63.8\% & 80.5\% & 38.6\% & 41.7\% \\
\textbf{DuATM~\cite{DBLP:journals/corr/abs-1803-09937}} & 76.6\% & 91.4\% & 64.6\% & 81.8\% & -- & -- \\
\textbf{DeepPerson~\cite{DBLP:conf/icb/JinWLL17}} & 79.6\% & 92.3\% & 64.8\% & 80.9\% & -- & -- \\
\textbf{Fusion~\cite{DBLP:journals/corr/abs-1803-10630}} & 79.1\% & 92.1\% & 64.8\% & 80.4\% & -- & -- \\
\textbf{SphereReID~\cite{DBLP:journals/corr/abs-1807-00537}} & 83.6\% & 94.4\% & 68.5\% & 83.9\% & -- & -- \\ \hline
\textbf{PL-Net~\cite{DBLP:journals/corr/YaoZZLT17}} & 69.3\% & 88.2\% & -- & -- & -- & -- \\
\textbf{PCB~\cite{DBLP:conf/eccv/SunZYTW18}} & 77.4\% & 92.3\% & 66.1\% & 81.7\% & 53.2\% & 59.7\% \\
\textbf{PCB+RPP~\cite{DBLP:conf/eccv/SunZYTW18}} & 81.6\% & 93.8\% & 69.2\% & 83.3\% & 57.5\% & 63.7\% \\
\textbf{MGN~\cite{DBLP:conf/mm/WangYCLZ18}} & 86.9\% & 95.7\% & 78.4\% & 88.7\% & 66.0\% & 66.8\% \\ \hline
\textbf{HBMCN} & 85.7\% & 94.4\% & 74.6\% & 85.7\% & 69.0\% & 73.8\% \\ \hline
\end{tabular}
\end{table*}

\textbf{Results on Market}: From Table \ref{tab_state-of-the-arts}, our HBMCN achieves 94.4\% in Rank-1 and 85.7\% in mAP, which surpasses most of the existing methods. Among these methods, we note that the IDE model~\cite{DBLP:journals/corr/ZhengYH16} is a widely used baseline for deep re-ID systems. Our results exceed IDE by +18.8\% in Rank-1 and +35.0\% in mAP. SphereReID~\cite{DBLP:journals/corr/abs-1807-00537} is one of the newest methods without local part information, achieving 94.4\% in Rank-1 and 83.6\% in mAP. Our method exceeds it by +2.1\% in mAP.

Comparing with part-based models, a gap is observed between our HBMCN and MGN, mainly because these methods use local fine-grained information and effective triplet loss to learn more discriminative features. However, our model has more advantages than PL-Net, PCB and PCB+RPP.

\textbf{Results on Duke}: The results achieved by our HBMCN show an excellent performance. HBMCN achieves Rank-1=85.7\% and mAP=74.6\%, outperforming the SphereReID with +1.8\% in Rank-1 and +6.1\% in mAP. Compared with PCB+RPP, HBMCN shows a better performance, exceeding by a large margin (+2.4\% in Rank-1 and +5.4\% in mAP).

\textbf{Results on CUHK}: As presented in Table \ref{tab_state-of-the-arts}, our HBMCN achieves a state-of-the-art result of Rank-1=73.8\% and mAP=69.0\% on the detected setting, which outperforms the part-based model, MGN, by +7.0\% in Rank-1 and +3.0\% in mAP. Note that CUHK provides both detected images using the pedestrian detector (detected setting) and human-labeled bounding boxes (labeled setting). The detected setting is harder than the labeled setting, as the detection failure has a great influence on re-ID.

\subsection{Component Analysis}
\subsubsection{Heterogeneous Branch}
\begin{table*}[!t]
\centering
\caption{Heterogeneous Branch Analysis on Market.}
\label{tab_heterogeneous}
\begin{tabular}{|l|c|c|c|c|c|}
\hline
\textbf{Methods} & \textbf{mAP} & \textbf{Rank1} & \textbf{Rank5} & \textbf{Rank10} & \textbf{Rank20} \\ \hline
{\textbf{Baseline}} & 78.67\% & 91.83\% & 96.91\% & 97.86\% & 98.60\% \\
{\textbf{+Res-Branch}} & 81.76\% & 92.49\% & 97.33\% & 98.19\% & 98.90\% \\
{\textbf{+SE-Res-Branch}} & 82.73\% & 92.84\% & 97.77\% & 98.75\% & 99.23\% \\ \hline
{\textbf{Single Res-Branch}} & 79.18\% & 91.30\% & 97.06\% & 98.22\% & 98.75\% \\
{\textbf{Single SE-Res-Branch}} & 76.94\% & 90.86\% & 97.06\% & 98.19\% & 98.96\% \\ \hline
\end{tabular}
\end{table*}

In order to compare the existing network with isomorphic branches, we build a network with two isomorphic branches by copying the structure and parameters from \emph{Res-Branch}. They are compared with the simplified version of our proposed HBMCN with only one objective at the end of each branch. The results are presented at the top of Table \ref{tab_heterogeneous}, where \emph{Baseline} is ResNet-50 with one \emph{Reduction} and one \emph{Objective} at the \emph{res\_conv5c}. \emph{+Res-Branch} builds a new branch by copying the \emph{res\_conv4x} and \emph{res\_conv5x} layers of ResNet-50. The feature is obtained by concatenating the features extracted from the layer \emph{res\_conv5c} in each branch. \emph{+SE-Res-Branch} is similar to \emph{+Res-Branch}, where the new branch is \emph{SE-Res-Branch}. Both \emph{+Res-Branch} and \emph{+SE-Res-Branch} have one \emph{Reduction} and one \emph{Objective} at the end of each branch. The mAP and Rank-1 performance are evaluated on the Market dataset.

From Table \ref{tab_heterogeneous}, we can see that \emph{Baseline} achieves 91.83\% in Rank-1, which is a very strong baseline. Comparing with \emph{Baseline}, \emph{+Res-Branch} improves the performance with +0.66\% in Rank-1 and +3.9\% in mAP, while \emph{+SE-Res-Branch} improves the performance with +1.01\% in Rank-1 and +4.06\% in mAP. From these results, we can conclude that two branches can boost the performance greatly. Meanwhile, \emph{SE-Res-Branch} has a better performance compared to \emph{Res-Branch}. This is because the Squeeze-and-Excitation block can focus its attention on the more important feature maps.

To further analyze the contribution of each branch, we extract the features from \emph{Res-Branch} and \emph{SE-Res-Branch} of \emph{+SE-Res-Branch} and evaluate the performance. The results are shown in the lower part of Table \ref{tab_heterogeneous}, where \emph{Single Res-Branch} and \emph{Single SE-Res-Branch} are the results of two corresponding branches.

It can be observed from the lower part of Table \ref{tab_heterogeneous} that \emph{Single Res-Branch} achieves better mAP and Rank-5/10/20 than \emph{Baseline}, as \emph{SE-Res-Branch} helps \emph{Res-Branch} in training. Furthermore, \emph{Single Res-Branch} also outperforms \emph{Single SE-Res-Branch} with +0.44\% n Rank-1 and +2.24\% in mAP. This is because \emph{SE-Res-Branch} is initialized with random parameters while \emph{Res-Branch} is initialized with the ImageNet pre-trained parameters.

\subsubsection{Multi-level Classification}
\begin{table*}[!t]
\centering
\caption{Multi-level Classification Analysis on Market.}
\label{tab_multi_level_cls}
\begin{tabular}{|l|c|c|c|c|c|}
\hline
\textbf{Methods} & \textbf{mAP} & \textbf{Rank1} & \textbf{Rank5} & \textbf{Rank10} & \textbf{Rank20} \\ \hline
{\textbf{Baseline+2Level}} & 81.93\% & 92.31\% & 97.06\% & 98.28\% & 98.96\% \\
{\textbf{+Res-Branch+2Level}} & 82.67\% & 93.08\% & 97.30\% & 98.16\% & 98.96\% \\
{\textbf{+SE-Res-Branch+2Level}} & 84.36\% & 94.09\% & 97.71\% & 98.75\% & 99.14\% \\ \hline
{\textbf{HBMCN}} & 85.68\% & 94.42\% & 98.01\% & 98.93\% & 99.17\% \\ \hline
\end{tabular}
\end{table*}

To verify the contribution of the multi-level classification, we simply set two classification objectives at \emph{res\_conv4f} and \emph{res\_conv5c} in the \emph{Res-Branch}. \emph{SE-Res-Branch} also has two objectives at the corresponding layers \emph{SE-Res4f} and \emph{SE-Res5c}. The results are showed in Table \ref{tab_multi_level_cls}.

From Table \ref{tab_multi_level_cls}, it can be seen that \emph{Baseline+2Level} achieves 92.31\% in Rank-1 and 81.93\% in mAP, with an obvious improvement of 0.48\% in Rank-1 and 3.26\% in mAP. Similar results are achieved by \emph{+Res-Branch+2Level} and \emph{+SE-Res-Branch+2Level}. When multi-level classification objectives are set at layers of \emph{res\_conv4f}, \emph{res\_conv5a}, \emph{res\_conv5b}, \emph{res\_conv5c}, \emph{SE-Res4f}, \emph{SE-Res5a}, \emph{SE-Res5b} and \emph{SE-Res5c}, HBMCN achieves 94.41\% in Rank-1 and 85.68\% in mAP.

Tables \ref{tab_heterogeneous} and \ref{tab_multi_level_cls} show that both \emph{SE-Res-Branch} and multi-level classification improve performance. Comparing with part-based models, we can conclude that multiple branches also have a great ability to improve the effectiveness, without local part information.

\section{Conclusion}
In this paper, we proposed a heterogeneous branch and multi-level classification network based on ResNet-50, and explored its priority. A new \emph{SE-Res-Branch} is proposed in order to enhance the capability. A new multi-level classification joint objective function is proposed to learn the parameters of HBMCN. Multi-level features are extracted and concatenated as the representation of a pedestrian. On three large-scale person re-ID datasets, experiments were performed to evaluate the effectiveness of our HBMCN. According to the evaluation results, we can conclude that HBMCN is an effective method for person re-ID. Extensive experimental analysis shows that the result achieved with the heterogeneous branch is better than that achieved with the isomorphic branch. In addition the multi-level classification provides more discriminative features compared to single-level classification. As a result, HBMCN provides substantial further improvements in person re-ID tasks. However, as with existing multi-branch networks, HBMCN has more parameters and a greater computation cost than the original ResNet-50.

In the future, the following directions can potentially be explored. The most important is the determination of why multiple branches improve performance. According to the work of~\cite{DBLP:journals/corr/KeskarMNST16}, the generalization gap between training and testing is related to the flatness of the minimum of the objective. A flat minimum can achieve good generalization, while a sharp minimum can worsen the generalization. As a result, if there are multiple minima, the flatness of the minima of the objectives can be expanded. For a classic CNN network, since the task is always a non-convex optimization problem, the solution is just one of the multiple minima. However, HBMCN has multiple objectives with heterogeneous branches, thus it can reach multiple minima and form a large scope of flat minima.

Furthermore, according to the work of~\cite{DBLP:conf/eccv/SunZYTW18,DBLP:conf/mm/WangYCLZ18}, a part-based model can represent local information, thus it can be combined with our work. In addition, it is also possible to build branches based on stronger blocks, such as a Dense block, in order to promote the performance.

\bibliographystyle{elsarticle-num}
\bibliography{HBMCN2019}

\end{document}